%% file: sample-base.tex
\documentclass[sigconf,natbib=true,anonymous=false]{acmart}
\makeatletter
\def\@ACM@checkaffil{
    \if@ACM@instpresent\else
    \ClassWarningNoLine{\@classname}{No institution present for an affiliation}%
    \fi
    \if@ACM@citypresent\else
    \ClassWarningNoLine{\@classname}{No city present for an affiliation}%
    \fi
    \if@ACM@countrypresent\else
        \ClassWarningNoLine{\@classname}{No country present for an affiliation}%
    \fi
}
\makeatother
\AtBeginDocument{%
  \providecommand\BibTeX{{%
    \normalfont B\kern-0.5em{\scshape i\kern-0.25em b}\kern-0.8em\TeX}}}

\copyrightyear{2023}
\acmYear{2023}
\setcopyright{acmlicensed}\acmConference[SIGIR '23]{Proceedings of the 46th International ACM SIGIR Conference on Research and Development in Information Retrieval}{July 23--27, 2023}{Taipei, Taiwan}
\acmBooktitle{Proceedings of the 46th International ACM SIGIR Conference on Research and Development in Information Retrieval (SIGIR '23), July 23--27, 2023, Taipei, Taiwan}
\acmPrice{15.00}
\acmDOI{10.1145/3539618.3591690}
\acmISBN{978-1-4503-9408-6/23/07}

\usepackage{caption}
\usepackage[labelformat=simple]{subcaption}
\usepackage{stfloats}
\usepackage{algorithmic}
\usepackage{graphicx}

\usepackage{amsmath}
\usepackage{amsfonts}
\usepackage{mathrsfs}
\usepackage{multirow}
\usepackage{booktabs}
\usepackage{pifont}
\usepackage{amssymb}
\usepackage{color}

\newcommand{\ie}{\emph{i.e.,}~}
\newcommand{\etal}{\emph{et al.}~}

\usepackage{color}

\newcommand{\myeq}[1]{Eq.\ref{#1}}
\begin{document}

\title{
From Region to Patch: 
Attribute-Aware Foreground-Background Contrastive Learning for Fine-Grained Fashion Retrieval}

\author{Jianfeng~Dong}
\affiliation{%
  \institution{Zhejiang Gongshang University}
  \institution{Zhejiang Key Lab of E-Commerce}
  }
\email{dongjf24@gmail.com}

\author{Xiaoman~Peng}
\affiliation{%
  \institution{Zhejiang Gongshang University}
  }
\email{pengxiaoman1999@gmail.com}

\author{Zhe~Ma}
\affiliation{%
  \institution{Zhejiang University}
  }
\email{mz.rs@zju.edu.cn}

\author{Daizong~Liu}
\affiliation{%
  \institution{Peking University}
  }
\email{dzliu@stu.pku.edu.cn}

\author{Xiaoye~Qu}
\affiliation{%
  \institution{Huazhong University of Science and Technology}
  }
\email{xiaoye@hust.edu.cn}

\author{Xun~Yang}
\affiliation{%
  \institution{University of Science and Technology of China}
  }
\email{hfutyangxun@gmail.com}

\author{Jixiang~Zhu}
\affiliation{%
  \institution{Zhejiang Gongshang University}
  }
\email{zhuzhu0111@163.com}

\author{Baolong~Liu}
\authornote{Corresponding author.}
\affiliation{%
  \institution{Zhejiang Gongshang University}
  \institution{Zhejiang Key Lab of E-Commerce}
  }
\email{liubaolongx@gmail.com}

\renewcommand{\shortauthors}{Jianfeng Dong et al.}

\begin{abstract}
Attribute-specific fashion retrieval (ASFR) is a challenging information retrieval task, which has attracted increasing attention in recent years. Different from traditional fashion retrieval which mainly focuses on optimizing holistic similarity, the ASFR task concentrates on attribute-specific similarity, resulting in more fine-grained and interpretable retrieval results. As the attribute-specific similarity typically corresponds to the specific subtle regions of images, we propose a \textit{Region-to-Patch Framework (RPF)} that consists of a region-aware branch and a patch-aware branch to extract fine-grained attribute-related visual features for precise retrieval in a coarse-to-fine manner. In particular, the region-aware branch is first to be utilized to locate the potential regions related to the semantic of the given attribute. Then, considering that the located region is coarse and still contains the background visual contents, the patch-aware branch is proposed to capture patch-wise attribute-related details from the previous amplified region. Such a hybrid architecture strikes a proper balance between region localization and feature extraction.  Besides, different from previous works that solely focus on discriminating the attribute-relevant foreground visual features, we argue that the attribute-irrelevant background features are also crucial for distinguishing the detailed visual contexts in a contrastive manner. Therefore, a novel \textit{E-InfoNCE} loss based on the foreground and background representations is further proposed to improve the discrimination of attribute-specific representation. Extensive experiments on three datasets demonstrate the effectiveness of our proposed framework, and also show a decent generalization of our RPF on out-of-domain fashion images. Our source code is available at \url{https://github.com/HuiGuanLab/RPF}.
\end{abstract}


\maketitle

\input{intro}

\input{relwork}

\input{method}

\input{eval}

\section{Conclusion}
This paper has contributed a novel Region-to-Patch Framework (RPF), which consists of a region-aware branch and a patch-aware branch to extract attribute-related features from the coarse-grained level to the fine-grained level, to address the challenging attribute-specific fashion retrieval task. 
With the further proposed foreground-background contrastive learning paradigm, by mining suitable positive and negative samples, the discrimination of attribute-related and attribute-unrelated representations can be improved for better retrieval.
Extensive experiments on three datasets demonstrate the effectiveness of our model for fine-grained attribute-specific fashion retrieval. Besides, our proposed RPF also achieves good generalization on out-of-domain data.

\section{ACKNOWLEDGMENTS}
This work was supported by the \textit{Pioneer} and \textit{Leading Goose} R\&D Program of Zhejiang (No.2023C01212), the Public Welfare Technology Research Project of Zhejiang Province (LGF21F020010),
Young Elite Scientists Sponsorship Program by CAST (2022QNRC001),  the NSFC (61976188), the Open Project of Key Laboratory of Public Security Information Application Based on Big-Data Architecture, Ministry of Public Security (2021DSJSYS001),
the Open Projects Program of the State Key Laboratory of Multimodal Artificial Intelligence Systems, and the Fundamental Research Funds for the Provincial Universities of Zhejiang.


\bibliographystyle{ACM-Reference-Format}
\bibliography{sample-base}

\end{document}

%% file: intro.tex
\section{Introduction}

Fashion retrieval is one of the important tasks in the information retrieval community \cite{yang2020tree,dong2022partially,zheng2023progressive,liu2021context,liu2020jointly,shen2022semicon}, which has a broad application in various e-commerce platforms, including fashion item recommendation~\cite{yang2019transnfcm,deldjoo2022review,de2022disentangling}, fashion trend forecasting~\cite{ma2020knowledge,al2017fashion,mall2019geostyle}, plagiarized fashion item detection~\cite{lang2020plagiarism,ma2020fine,kim2020plagiarism}, fashion matching~\cite{dong2020fashion,yang2019interpretable,ma2021hierarchical}, and so on.
The traditional fashion retrieval task~\cite{zhang2018visual,kang2019complete} aims to retrieve similar fashion items with the query image holistically (as exemplified in Figure \ref{fig:intro}), which typically measures the overall similarity among fashion items in a learned common embedding space.
Different from traditional fashion retrieval, a new but challenging Attribute-Specific Fashion Retrieval (ASFR) task has been proposed for more fine-grained fashion retrieval~\cite{veit2017csn,ma2020fine}.
As shown in the second row of Figure \ref{fig:intro}, given a query image and a certain attribute, such as \textit{neckline design}, ASFR aims to retrieve fashion images containing the subtle details of the same attribute value, \ie \textit{V Neck}, with the query image. 
Such retrieval paradigm has potential value in many fashion-related information retrieval applications, such as fine-grained fashion retrieval where one could search for fashion items with specific designs, and fashion copyright protection where one would like to retrieve plagiarized items that plagiarized certain parts of the original one.
In this paper, we focus on addressing the challenging ASFR task.

\input{eval/intro}

The key of the ASFR task is how to accurately perceive the attribute-related patterns of images and compute the fine-grained fashion similarity in terms of the specific attribute.
However, this is challenging as the attribute-related patterns typically only cover a small part of the image instead of the whole image.
Recently, significant efforts have been made for learning fine-grained fashion similarity for ASFR~\cite{veit2017csn,yan2021learning,wan2022learning,yan2022attribute}.
As a pioneer work of ASFR, Veit \etal \cite{veit2017csn} first learn an overall embedding of the whole image, and then obtain the attribute-aware features by employing fixed masks with respect to the specified attribute on the overall embedding.
After that, a number of follow-up works prefer to utilize attention mechanisms with the guidance of the attribute to capture the attribute-related patterns~\cite{ma2020fine,yan2022attribute}.
For instance, Ma \etal \cite{ma2020fine} propose to first locate the attribute-related regions by an attribute-aware spatial attention, then an attribute-aware channel attention is further used to derive the attribute-related features.
Yan \etal \cite{yan2022attribute} propose to repeatedly employ attention to extract fine-grained features step-by-step.
Similarly, Dong \etal \cite{dong2021fine} also conduct attention mechanisms repeatedly, and they devise a two-branch network consisting of a global branch and a local branch.
However, they utilize the same architecture for two branches,
which not only limits the complementarity of the two-branch network, but also solely captures the coarse region-aware attribute-related visual contexts, failing to distinguish the subtle details of some challenging attribute.
However, only capturing the region-level attribute-related context is coarse and not enough, since more distinguishable attribute information is subtle and the region contexts still contain many attribute-irrelevant visual contents.
Therefore, in order to strengthen the complementarity between multiple branches while capturing more fine-grained visual details related to the attribute, it is necessary to design a representation learning network with a distinct and gradual granularity for each branch, while exploring the cross-branch consistency learning mechanism.

To this end, we propose a novel \textit{Region-to-Patch Framework (RPF)}, which consists of coarse-to-fine encoding branches to extract different granularities of features from region to patch level for capturing more fine-grained attribute-related visual details.
Our motivation is inspired by the natural image understanding of humans, where humans typically locate certain content in images by first glancing at the whole image and then searching the subtle contexts in part progressively.
Specifically,
our proposed RPF has two branches, a region-aware branch (coarse-level) and a patch-aware branch (fine-level), which explore the fine-grained features at different granularities. 
Given an image and an attribute, the region-aware branch first encodes both image and attribute, then attentively locates the relevant foreground region corresponding to the semantic of the attribute. 
Since the foreground (attribute-relevant) representation is relatively coarse and the attended region still contains attribute-irrelevant information, a patch-aware branch is further developed to explore more fine-grained attribute-related patch-aware contexts based on the previous foreground region.
In particular, the foreground region from the region-aware branch is amplified and divided into multiple non-overlapping patches, which are fed into the patch-aware branch for learning to focus on the interested patches of finer granularity under the guidance of the attribute.

Besides, to ensure both region-aware and patch-aware branches learn properly, we further propose a foreground-background contrastive learning strategy.
To be specific, since the attribute-relevant foreground features of the same attribute are expected to be as close as possible while the attribute-irrelevant ones are expected to be far away, we develop an intra-branch contrastive loss to discriminate their representations.
In addition, since we obtain the coarse and fine-grained attribute-specific features from the two branches, we aim to constrain their semantic consistency as they aim to obtain the same semantic pattern from images. Therefore, we also devise an inter-branch contrastive loss to encourage the two branches to have a consistent alignment that assigns similar representations to similar samples corresponding to the same attribute.
Moreover, we propose E-InfoNCE as the inter-branch contrastive loss, which enhances positive samples with attribute-relevant foregrounds and  negative samples with attribute-irrelevant backgrounds.
In summary, the contributions of this work are summarized as follows:
\begin{itemize}
\item We propose a novel \textit{Region to Patch Framework} which consists of two coarse-to-fine encoding branches, including a region-aware branch and a patch-aware branch, which progressively extracts different granularities of attribute-related visual features from region to patch level.
\item In order to effectively train the above two-branch network, we propose a foreground-background contrastive learning paradigm that not only learns the consistent attribute-related visual semantics among two branches but also discriminates these semantics of different attributes. 
Moreover, we devise a new contrastive loss, E-InfoNCE, which enhances the common contrastive loss InfoNCE~\cite{oord2018representation} by mining more positive and negative samples.
\item Extensive experiments on FashionAI \cite{zou2019fashionai}, DARN \cite{huang2015cross} and DeepFashion \cite{liu2016deepfashion}  demonstrate the effectiveness and generalization of our proposed method, and 
we establish a new state-of-the-art for ASFR on all of these datasets.
\end{itemize}

%% file: eval/intro.tex
\begin{figure}[!t]
\centering
\includegraphics[width=.95\linewidth]{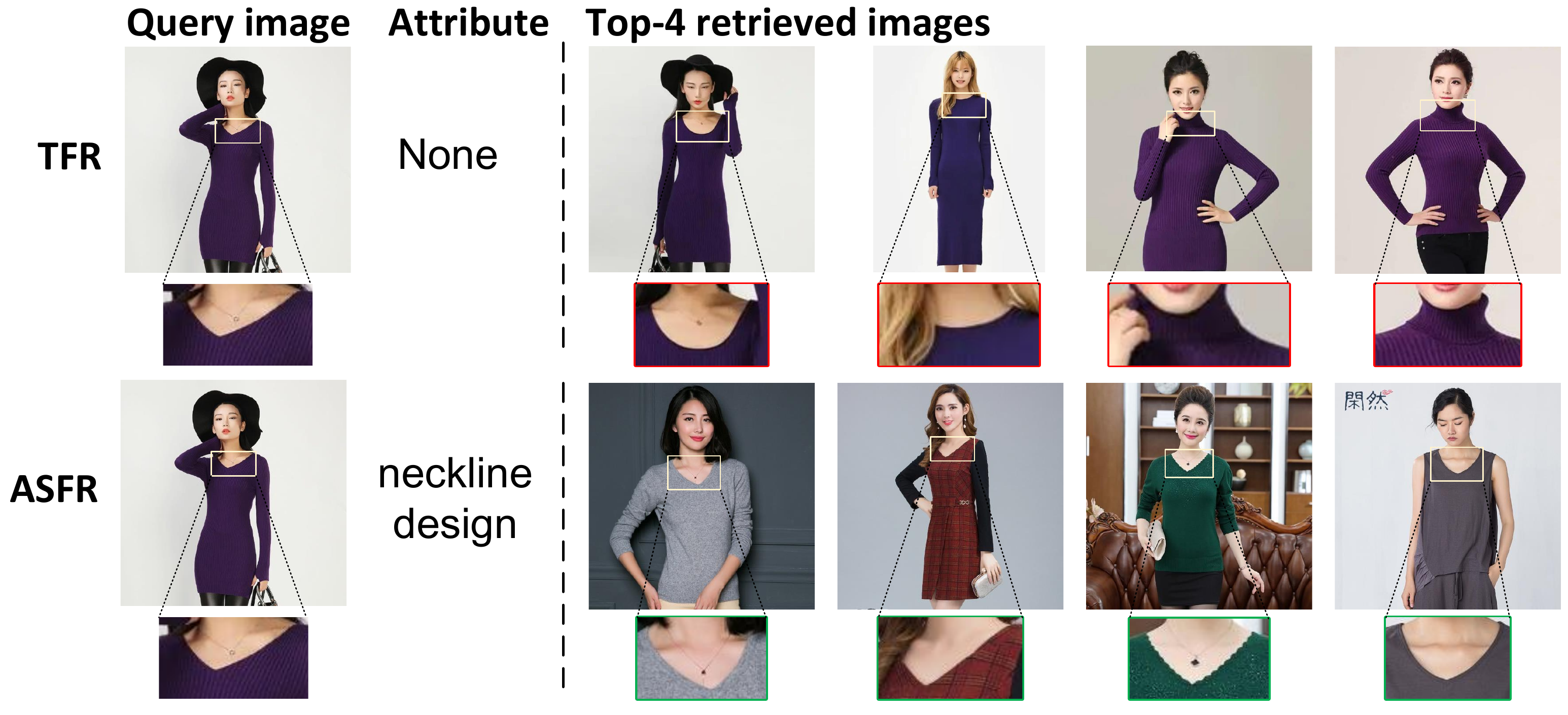}
\vspace{-4mm}
\caption{The comparison between the traditional fashion retrieval and the attribute-specific fashion retrieval. The former mainly focuses on holistic similarity, while the latter solely focuses on more fine-grained similarity with respect to the given attribute.}
\label{fig:intro}
\vspace{-5mm}
\end{figure}

%% file: relwork.tex
\section{Related Work}\label{sec:relwork}

\subsection{Traditional Fashion Retrieval}
Fashion retrieval is a long-standing task in the fashion community, and has achieved great progress in recent years. It mainly can be categorized into following subtasks, \ie in-shop fashion retrieval~\cite{luo2019spatial,kinli2019fashion,wang2017clothing}, street-to-shop fashion retrieval~\cite{alirezazadeh2022deep,kuang2019fashion,morelli2021fashionsearch++} and fashion compatibility prediction~\cite{kim2021self,mishra2021effectively,lin2020fashion}, and so on.
For in-shop fashion retrieval, it aims to retrieve images containing identical or similar clothes with the query image, from a gallery of candidate images.
Different from the in-shop fashion retrieval that assumes that the query and candidate image are from the same domain, street-to-shop fashion retrieval allows images to be from different domains and is more challenging.
The above two tasks aim to retrieve fashion items of the same category, the fashion compatibility prediction task learns the visual compatibility or functional complementarity between fashion items, typically between different categories.
Different from the above tasks focus on the overall similarity between fashion items, this work aims to learn attribute-aware similarity, a more fine-grained similarity paradigm.

\input{eval/framework}

\subsection{Attribute-Specific Fashion Retrieval}
Different from the traditional fashion retrieval task, ASFR needs to extract the features of the local region which is related to a specified fashion attribute, rather than the whole image features.
ASFR is more challenging and has attracted increasing attention recently~\cite{ma2020fine,veit2017csn,yan2022attribute}.
The majority of works tend to utilize attention mechanisms with the guidance of the attribute to capture the attribute-aware representation~\cite{yan2022attribute,yan2021learning,wan2022learning}.
For instance, Ma \etal \cite{ma2020fine} propose a model named ASEN, which first locates the attribute-related region by an attribute-aware spatial attention (ASA), and further extracts fine-grained features by an attribute-aware channel attention (ACA).
On the basis of ASEN, Yan \etal \cite{yan2021learning} additionally propose a hierarchical attribute embedding module to enhance the relationship between attributes.
Wan \etal \cite{wan2022learning} applies ASA and ACA from ASEN in a parallel manner, then fuses features extracted by both attention modules as the final feature representations. 
Recently, Yan \etal \cite{yan2022attribute} repeatedly employ attention to extract fine-grained features step-by-step, where a general framework of conducting multiple attentions iteratively is proposed.
Different from the works that utilize a one-branch network, Dong \etal \cite{dong2021fine} propose a two-branch network of the same architecture for each branch. They extract the fine-grained features from the global and local perspectives respectively.
Our proposed model is also a two-branch network, but we devise a coarse-to-fine network with a distinct granularity for each branch. 
Besides, the previous works all discard attribute-unrelated patterns, in this work we exploit the attribute-unrelated patterns and found they are beneficial for ASFR.

%% file: eval/framework.tex
\begin{figure*}[!t]
\centering
\includegraphics[width=1.0\textwidth]{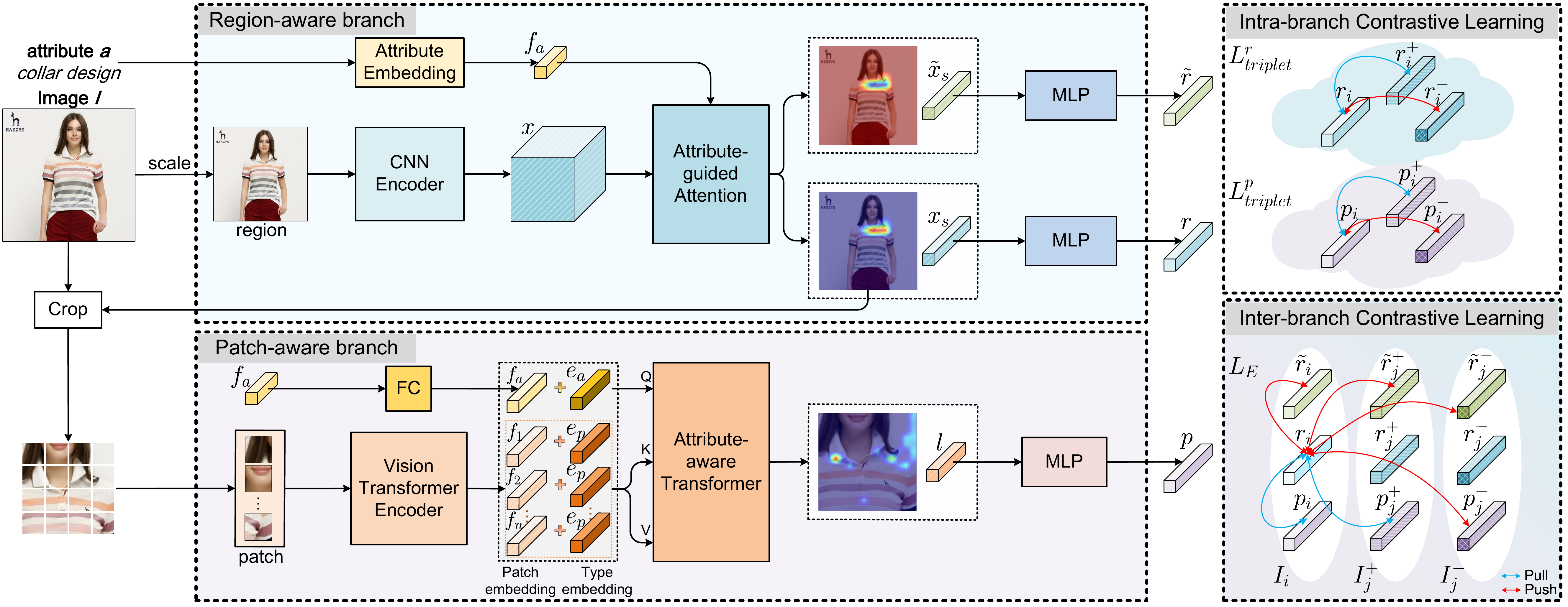}
\vspace{-8mm}
\caption{The framework of our proposed RPF that consists of a CNN-based region-aware branch (coarse-level) and a Transformer-based patch-aware branch (fine-level).
}
\label{fig:framework}
\end{figure*}

%% file: method.tex
\section{The Proposed Method}
\subsection{Overview of Region-to-Patch Framework}

The ASFR task aims to finely retrieve fashion images sharing the same attribute-specific knowledge.
To this end, we propose a novel \textit{Region-to-Patch Framework (RPF)} as illustrated in Figure \ref{fig:framework}, which consists of coarse-to-fine encoding branches to extract different granularities of features from region to patch level for precise retrieval.
Specifically,
given an image $I$ and an attribute $a$, the region-aware branch (coarse-level) first encodes both of them and attentively locates the attribute-relevant foreground region corresponding to the semantic of attribute $a$. 
Then, it disentangles the mixed representations of the entire image into two contradictory parts $r$ and $\widetilde{r}$, representing the attribute-related foreground and background regions, respectively.
Since this foreground representation is relatively coarse and still contains attribute-irrelevant visual information, a  patch-aware branch (fine-level) is developed to explore more fine-grained attribute-related contexts based on the previous foreground region.
In particular, the foreground region from the region-aware branch is amplified and divided into multiple non-overlapping patches which are fed into the current branch as input. 
The patches interact with each other to learn their self-relation contexts via a vision Transformer encoder, and an attribute-aware Transformer
module is further devised to look for the interesting patches under the guidance of the attribute, resulting in a more fine-grained representation $p$.
A foreground-background contrastive learning is further employed to discriminate the representation learning of the attribute-specific foreground-background image contexts.
In what follows, we describe each component in detail.

\subsection{Region-aware Branch}
\subsubsection*{\bf Learning to focus on the attribute-related foreground region across the entire image}
The region-aware branch is supposed to be capable of adaptively localizing the attribute-relevant foreground region from the original full image $I$ and learning the corresponding region-aware representation. 
Instead of solely extracting foreground features of the image while discarding the attribute-independent background components as previous works \cite{veit2017csn,yan2021learning,wan2022learning}, we argue that these discarded background features can serve as the natural negative samples for the foreground features in a contrastive manner, promoting to learn more robust and discriminative representations.
Therefore, in this region-aware branch, we devise an attention module to disentangle the entire image representation into two complementary representations, \textit{i.e.}, foreground and background representations, and introduce a foreground-background representation contrastive learning strategy in Section \ref{ssec:cl}.

\subsubsection{Foreground Representation}
Given an image and an attribute, the attribute-relevant foreground representation is obtained by an attribute-guided attention module.
Specifically, the input image $I$ is first fed into a CNN backbone encoder, \ie ResNet50, to obtain the feature map of the full image, denoted as $x\in\mathbb{R}^{c\times h\times w}$, while the input attribute $a$ is encoded by an attribute embedding module into an attribute embedding vector which is denoted as $f_a\in \mathbb{R}^{c_a}$.
Considering that the image content related to the given attribute generally appears in a certain region, we aim to focus on learning the attribute-related feature of this specific region instead of the full image. 
As the image and the attribute are of different modalities, we first project them into a joint latent space.
Concretely, we separately utilize a $1\times1$ convolution layer and an FC layer to project the image feature and attribute embedding into a joint latent space, followed by a nonlinear activation function $\tanh$, obtaining the projected feature map and attribute embedding as $x_c\in\mathbb{R}^{c_m\times hw}$ and ${f_{ac}}\in \mathbb{R}^{1\times{c_m}}$.
Then an attribute-guided attention module is exploited to generate an attributed-guided foreground attention map $\alpha\in \mathbb{R}^{h\times w}$ by measuring the similarities between the feature map $x_c$ and attribute embedding $ {f_{ac}}$ at each spatial position as:
\begin{equation}
\alpha = \text{softmax}( {f_{ac}} \cdot x_c).
\end{equation}
This probabilistic attention map $\alpha$ is then used to disentangle the representations of the full image into foreground (attribute-relevant) and background (attribute-irrelevant) representations.

Concretely, the foreground representation $ {x}_s\in \mathbb{R}^c$ is calculated by a weighted summation of $x$ according to the attention map $\alpha$ over the spatial dimension as: 
$ {x}_s = \sum_{i}^{h\times w} \alpha_{i}  {x}_i$,
where $ {x}_i$ is the $i$-th channel-aware feature vector of the feature map $x$.
Then, a Multilayer Perceptrons (MLP) layer with layer normalization (LN) and residual connections are employed to obtain the final foreground representation $r\in \mathbb{R}^{c_o}$ as follow:
\begin{equation}
 r = LN(W_2(\text{relu}(W_1(LN( {x}_s)))) + {x}_s) ,
\end{equation}
where $LN$ is layer normalization, $W_1$ and $W_2$ are trainable transformation weights.

\subsubsection{Background Representation}
In order to obtain the background representation that is irrelevant to the given attribute, we first generate a background attention map where high values are given for the irrelevant regions while low values are for the relevant ones.
As the foreground and background representations are dependent, we generate the background attention map based on the foreground attention map instead of learning a new one. 
Specifically, we simply inverse the foreground attention map $\alpha$, and further utilize a linear normalization to normalize the inversed attention map. Formally, the background attention map is computed as:
\begin{equation}\label{alphahat}
\widetilde{\alpha} = \frac{1-\alpha}{\sum_{i}^{h\times w} (1- \alpha_{i})}.
\end{equation}
With the background attention map, the background representation $ {\widetilde{x}}_s\in \mathbb{R}^c$ is calculated as:
$ {\widetilde{x}}_s = \sum_{i}^{h\times w} \widetilde{\alpha}_{i}  {x}_i$.
Similar to the foreground representation, the same MLP is further utilized, and the final background representation $ {\widetilde{r}}\in \mathbb{R}^{c_o}$ is obtained as:
\begin{equation}
 {\widetilde{r}} = LN(W_2(\text{relu}(W_1(LN( {\widetilde{x}_s})))) +  {\widetilde{x}}_s).
\end{equation}

\subsection{Patch-aware Branch}
\subsubsection*{\bf Diving into more fine-grained patch context from the region}
As the attribute-relevant regions detected by the attribute-guided attention module are typically small, it prevents the region-aware branch from capturing the attribute-relevant information adequately.
To alleviate it, we introduce a patch-aware branch that takes the zoom-in attribute-relevant regions as the input and extracts features patch-wise in a more fine-grain manner.
For ease of reference, we refer to the attribute-relevant regions as the \textbf{region of interest} (RoI). The RoI is obtained by cropping from the full image according to the attention weights.
To extract patch-wise features, we first split the region into multiple smaller patches and then interact attribute with patches to determine whether each patch is related to the attribute semantics.

Considering ViT~\cite{2021An} is suitable to learn the representation of images in a patch form, here we choose it as the encoder of our patch-aware branch. Concretely, we first split the RoI into $16\times 16$ patches, which are regarded as image tokens analogous to word tokens in natural language processing. 
Pixels of each patch are flattened and linearly transformed to the hidden representation, subsequently added by position embeddings, resulting in a sequence of patch embeddings.
Then, the patch embeddings are fed into the ViT backbone which consists of a $L$-layer Transformer encoder to learn contextual embeddings for each patch and obtain a sequence of patch representations, denoted as:
$F=[ {f}_1,  {f}_2, ...,  {f}_n]$,
where $ {f}_i\in \mathbb{R}^D$ is the $i$-th patch token embedding of the sequence, and $n$ indicates the number of patches.

As the RoI is generated without explicit supervision,
it is inevitable that the RoI may contain a small quantity of noisy elements related to other attributes.
To alleviate it, we further propose an attribute-aware Transformer module to filter out these noisy representations under the guidance of the attribute.
As the attribute and RoI are two different modalities, we first learn two types of embeddings ${[ {e_a}, {e_p}]}$ to distinguish embeddings of different modalities. Then we respectively add homologous type embedding into attribute and patch embeddings to obtain type-enhanced embeddings. 
Specifically, the type-enhanced attribute embedding $ {f_a}'$ and patch embeddings $F'$ are respectively obtained as:
\begin{equation}
\begin{aligned}
 {f_a}' = \text{FC}( {f_a}) +  {e_a},  \quad  F'=[ {f}_1 +  {e_p},  {f}_2 +  {e_p}, ...,  {f}_n +  {e_p}],
\end{aligned}
\end{equation}
where FC is a fully connected layer to obtain the same feature dimension as the patch embedding.

After that, an attention module is further employed to weaken the impact of the attribute-irrelevant patches under the guidance of the attribute.
As for this attention module, we borrow the idea of multi-head self-attention module in Transformer~\cite{2021An}, where the input is first projected into queries, keys, and values, and the output is computed as a weighted sum of the values.
We adapt the multi-head self-attention module by taking the attribute as the query, and the patch embeddings as keys and values.
Specifically, for each attention head, the attribute embedding $ {f_a}'$ is first linearly projected as the query, while the patch embeddings $F'$ are linearly projected as keys and values respectively.
Then a scaled dot-product attention is further utilized to measure the correlation between the query vector and each key vector, which is used to select the attribute-related features of the patch embedding sequence $F'$ and aggregate them as the attribute-related representation of the RoI. 
For $i$-th attention head, the attentive representation $ {l_i}\in \mathbb{R}^d$ is obtained as:
\begin{equation}\label{lu}
 {l_i} = \text{softmax}(\dfrac{QK^\mathsf{T}}{\sqrt{d}})V,
\end{equation}
 where $Q= {f_a}'W^q_i$, $K={F'}W^k_i$, $V={F'}W^v_i$, and $W^q_i, ~W^k_i, ~W^v_i\in\mathbb{R}^{D\times d}$ are three projection matrices. 
 After jointly learning $h$ attention heads, we concatenate their outputs followed by an output projection layer.
 Formally, the output of our attention is:
 \begin{equation}
 {l} = \text{Concat}( {l_{1}},  {l_{2}},\dots,  {l_{h}})W_o,
\end{equation}
where $W_o\in \mathbb{R}^{hd\times D}$ is a the output projection matrix.

Following the traditional Transformer \cite{vaswani2017attention}, we also enhance $ {l}$ with a residual connection.
A mean pooling operation is conducted on the sequence of patch embeddings $F'$, thus it can be added to $ {l}$.
Additionally, an MLP with a residual connection and a layer normalization is further employed to obtain the  attribute-related image representation $ {p}\in \mathbb{R}^{c_o}$, which is the final output of the patch-aware branch.

\subsection{Foreground-Background Contrastive Learning}\label{ssec:cl}
With the region-aware branch and the patch-aware branch, we can obtain both coarse and fine-grained attribute-specific features for each image.
In order to improve the attribute-aware discrimination of these features, we train the whole network with our proposed foreground-background contrastive learning module, which consists of an intra-branch contrastive learning loss and an inter-branch contrastive learning loss.

\subsubsection{Intra-branch Contrastive Learning}
For each branch, we expect their corresponding output attribute-related foreground representations to be close for the input images with the same specific attribute value while being far away for the images with different attributes.
Take the \textit{sleeve length} attribute for example, the fashion images with \textit{short sleeves} attribute value should stay close to those which are also \textit{short sleeves}, but far away from the fashion images that exhibit \textit{long sleeves} in the learned feature space.
Therefore, we separately utilize two contrastive losses on the region-aware branch and the patch-aware branch.

To be specific, we first construct a minibatch of triplets $\mathcal{B} = \{(I_i, I_i^+, I_i^- | a)\}_{i=1}^N$, where the attribute value of image $I_i$ is the same as image $I^+_i$ but different from image $I^-_i$ in terms of attribute $a$, and $N$ denotes the batch size.
We instantiate the contrastive loss with triplet ranking loss as previous works~\cite{veit2017csn,ma2020fine,dong2022dual,wang2022cross}.
Formally, given a minibatch, the contrastive loss for the region-aware branch is:
\begin{equation}\label{eq:tri_g}
\mathcal{L}^{r}_{triplet}=\frac{1}{N}\sum_{i=1}^N \max(0,m - s( {r}_i, {r}_i^+)+s( {r}_i, {r}_i^-)),
\end{equation}
where $m$ denotes the margin constraint, $ {r}_i,  {r}_i^+,  {r}_i^-$ are the attribute-related foreground representations of $i$-th triplet $I_i, I_i^+, I_i^-$ obtained by the region-aware branch,
and $s(,)$ is cosine similarity function.

Similarly, the contrastive loss of the patch-aware branch is:
\begin{equation}\label{eq:tri_l}
\mathcal{L}^{p}_{triplet}=\frac{1}{N}\sum_{i=1}^N \max(0,m - s( {p}_i, {p}_i^+)+s( {p}_i, {p}_i^-)),
\end{equation}
where $ {p}_i,  {p}_i^+,  {p}_i^-$ are the attribute-related foreground representations of $i$-th triplet $I_i, I_i^+, I_i^-$ obtained by the patch-aware branch.

\subsubsection{Inter-branch Contrastive Learning}
Although the two branches extract attribute-aware features at different granularities, they aim to obtain the same semantic pattern from images.
Therefore, we devise an inter-branch contrastive loss to encourage the two branches to have a consistent alignment that assigns similar representations to similar samples corresponding to the same attribute.
To this end, inspired by contrastive learning in unsupervised representation learning~\cite{he2020moco,chen2020simclr,caron2020swav}, we aim to maximize the mutual information between the two branches.

\input{table-sota-fashionai}
\input{table-sota-darn}

Specifically, we regard the representations obtained from the two branches, \ie ${r}_i$ and ${p}_i$, as the two views of the input image with respect to the given attribute, and maximize the mutual information between the two views.
To achieve the objective of mutual information maximization, the straightforward way is to utilize an InfoNCE loss~\cite{oord2018representation} over the two views, that is:
\begin{equation}\label{eq:info}
\begin{aligned}
\mathcal{L}\!&=-\frac{1}{N}\sum_{i=1}^N \text{log}\!
\left(
\frac{\exp{( {r}_i\cdot{ {p}_i}/\tau)}}
{\mathcal{Z}
}\right), \\
\mathcal{Z} &= \exp{( {r}_i\cdot{ {p}_i}/\tau)}
+\sum\limits_{ {p}_j^- \in \mathcal{N}_i}\exp{( {r}_i\cdot{ {p}_j^-}/\tau)},
\end{aligned}
\end{equation}
where $\tau$ is a temperature factor, $\mathcal{N}_i$ denotes the negative sample set of different attribute values with $I_{i}$ in the mini-batch, and all the negative samples are obtained as the foreground representation from the patch-aware branch.

The above loss is constrained to learn the consistency of the attribute-relevant (foreground) features between the region-aware and patch-aware branches.
However, it is also crucial to discriminate the attribute-relevant and attribute-irrelevant (background) features for better discriminating the RoIs.
Therefore, we enhance the InfoNCE loss by including the foreground representation of the same attribute value as positive samples, and the background representation as negative samples. 
Formally, an enhanced infoNCE loss (E-infoNCE) is defined as:
\begin{equation}\label{eq:bec}
\begin{aligned}
\mathcal{L}_{E}\!&=\!\!-\frac{1}{N}\!\!\sum_{i=1}^N\!log
\left(\!\!
\frac{\exp{( {r}_i\!\cdot\!{ {p}_i}/\tau)} \!+\! \sum\limits_{{p}_j^+ \in \mathcal{P}} \exp{( {r}_i\!\cdot\!{ {p}_j^+}/\tau)}}
{\mathcal{Z}\!+\!\sum\limits_{{p}_j^+ \in \mathcal{P}}\exp{( {r}_i \!\cdot\!{ {p}_j^+}\!/\tau)} \!+\! \mu\!\!\!\!\sum\limits_{\widetilde{{r}_j}\in \mathcal{N}_k}\!\!\!\exp{( {r}_i\!\cdot\!{ \widetilde{{r}_j}}/\tau)}
}\!\!\right),
\end{aligned}
\end{equation}
where $\mathcal{P}$ indicates a positive set of foreground representations of the same attribute values with $I_{i}$, and $\mathcal{N}_k$ denotes a negative set of background representations of the same attribute with $I_{i}$ in the mini-batch.
Besides, $\mu$ is a scaling parameter for the background representation, and a larger $\mu$ indicates a stronger penalty on the similarity between the foreground and the background representations of the same attribute. 

\subsubsection{Training and Inference}

The total training loss is defined as:
\begin{equation}\label{eq:total}
\mathcal{L}_{total} = \mathcal{L}^{r}_{triplet} + \beta \mathcal{L}^{p}_{triplet} + \gamma \mathcal{L}_{E},
\end{equation}
where $\beta$ and $\gamma$ are hyper-parameters, which balance the importance of three losses.

During the inference stage, the similarity between query images and candidate images in terms of a certain attribute is computed as the weighted sum of  attribute-specific similarities of both branches, which is defined as:
\begin{equation}\label{lambda}
sim(I,I^*)=\lambda s( {r}, {r}^*) + (1-\lambda) s( {p}, {p}^*),
\end{equation}
where $\lambda$ is a weighting hyper-parameter.

%% file: table-sota-fashionai.tex
\begin{table*} [bp]
\renewcommand{\arraystretch}{1.}
\caption{Performance comparison on FashionAI. 
Methods are sorted in ascending order in terms of their overall performance.
}\label{tab:sota-fashinai}
\vspace{-3mm}
\centering 
\scalebox{0.87}{
\begin{tabular}{l*{9}{c}}
\toprule
\multirow{2}{*}{\textbf{Method}} & \multicolumn{8}{c}{\textbf{MAP for each attribute}} & \multirow{2}{*}{\textbf{overall MAP}} \\
\cmidrule(l){2-9}
& skirt length & sleeve length & coat length & pant length & collar design & lapel design & neckline design & neck design \\
\cmidrule(l){1-10}
Triplet baseline & 48.38 & 28.14 & 29.82 & 54.56 & 62.58 & 38.31 & 26.64 & 40.02 & 38.52 \\
CSN~\cite{veit2017csn}                 & 61.97 & 45.06 & 47.30 & 62.85 & 69.83 & 54.14 & 46.56 & 54.47 & 53.52 \\
$\text{ASEN}$~\cite{ma2020fine}     & 64.44 & 54.63 & 51.27 & 63.53 & 70.79 & 65.36 & 59.50 & 58.67 & 61.02\\
HAEN~\cite{yan2021learning}                & 64.13 & 55.52 & 56.41 & 72.31 & 73.32 & 69.22 & 62.41 & 59.80 & 64.13 \\
$\text{ASEN}^{++}$~\cite{dong2021fine}     & 66.34 & 57.53 & 55.51 & 68.77 & 72.94 & 66.95 & 66.81 & 67.01 & 64.31\\
AttnFashion~\cite{wan2022learning}         & 65.70 & 56.46 & 54.64 & 71.12 & 74.45 & 69.36 & 65.69 & 65.54 & 65.37 \\
ISLN~\cite{yan2022attribute} & 65.91 & 58.83 & 56.45 & 71.22 & 74.53 & 70.55 & 65.71 & 65.61 & 66.10\\
RPF (ours) & \textbf{66.75} & \textbf{67.86} & \textbf{59.65} & \textbf{73.23} & \textbf{75.72} & \textbf{73.18} & \textbf{74.40} & \textbf{75.01} & \textbf{70.11}\\
\bottomrule
\end{tabular}
}
\vspace{-1mm}
\end{table*}

%% file: table-sota-darn.tex
\begin{table*}[bp]
\renewcommand{\arraystretch}{1.}
\caption{Performance comparison on the DARN dataset.
\vspace{-3mm}
\label{tab:sota-darn}}
\centering 
\scalebox{0.75}{
\begin{tabular}{l*{10}{c}}
\toprule
\multirow{2}{*}{\textbf{Method}} & \multicolumn{9}{c}{\textbf{MAP for each attribute}} & \multirow{2}{*}{\textbf{overall MAP}} \\
\cmidrule(l){2-10}
 &clothes category&clothes button&clothes color&clothes length&clothes pattern&clothes shape & collar shape&sleeve length&sleeve shape \\
\cmidrule(l){1-11}
Triplet baseline & 23.59 & 38.07 & 16.83 & 39.77 & 49.56 & 47.00 & 23.43 & 68.49 & 56.48 & 40.14 \\
CSN~\cite{veit2017csn}                 & 34.10 & 44.32 & 47.38 & 53.68 & 54.09 & 56.32 & 31.82 & 78.05 & 58.76 & 50.86 \\
$\text{ASEN}$ ~\cite{ma2020fine}    & 36.69 & 46.96 & 51.35 & 56.47 & 54.49 & 60.02 & 34.18 & 80.11 & 60.04 & 53.31\\
HAEN  ~\cite{yan2021learning}              & 32.10 & 47.04 & 45.03 & 48.27 & 49.92 & 51.22 & 28.05 & 78.29 & 58.47 & 48.70 \\
AttnFashion ~\cite{wan2022learning}         & 34.94 & 48.56 & 48.14 & 54.47 & 52.65 & 56.36 & 32.32 & 82.63 & 60.77 & 52.32 \\
ISLN ~\cite{yan2022attribute}   & 38.84 & 51.26 & 52.67 & 56.55 & 53.85 & 58.34 & 36.64 & 82.74 & 61.28 & 54.68 \\
$\text{ASEN}^{++}$ ~\cite{dong2021fine}    & 40.15 & 50.42 & 53.78 & 60.38 & 57.39 & 59.88 & 37.65 & 83.91 & 60.70 & 55.94\\
RPF (ours)      & \textbf{45.18} & \textbf{54.92} & \textbf{55.08} & \textbf{63.51} & \textbf{57.04} & \textbf{63.54} & \textbf{41.20} & \textbf{86.95} & \textbf{62.43} & \textbf{58.80}\\
\bottomrule
\end{tabular}
}
\end{table*}

%% file: eval.tex
\input{table-sota-deepfashion}

\section{Evaluation}

\subsection{Experimental Setup} \label{ssec:exp-dataset}
\subsubsection{Datasets} \label{ssec:exp-dataset}
Following the previous works \cite{ma2020fine,yan2021learning,wan2022learning}, we conduct experiments on three fashion related datasets, \ie FashionAI \cite{zou2019fashionai}, DARN \cite{huang2015cross}, and DeepFashion \cite{liu2016deepfashion}.
These datasets were originally developed for fashion recognition or fashion retrieval, and have also been widely used for fine-grained attribute-specific fashion retrieval.

\textit{FashionAI} is a large-scale dataset that consists of 180,335 fashion images, where each image is annotated with a fine-grained attribute.
There are 8 attributes and each attribute has a corresponding list of attribute values.
For instance, the attribute \textit{sleeve length} has 9 attribute values, such as \textit{sleeveless}, \textit{short sleeves} and so on.
We use the data partition provided by \cite{ma2020fine}.

\textit{DARN} is originally constructed for fashion attribute prediction and cross-domain fashion image retrieval tasks, and has also been re-purposed for fine-grained attribute-specific fashion retrieval.
There are 214,619 images available, and each image is annotated with 9 attributes.
For data partition, following \cite{ma2020fine}, we utilize around 171k, 21k, 21k images for training, validation, and testing respectively.

\textit{DeepFashion} is a classical fashion related dataset, which has been commonly used for fashion category and attribute prediction, in-shop clothes retrieval, fashion landmark detection, and street-to-shop clothes retrieval. It has been reconstructed for attribute-specific fashion retrieval by \cite{ma2020fine}.
The dataset contains 289,222 images, 6 attributes and 1,050 attribute values.
For a fair comparison, we follow the data partition of \cite{ma2020fine}.

\subsubsection{Performance Metric}
Following the previous works \cite{ma2020fine,yan2021learning,wan2022learning}, we use the Mean Average Precision (MAP) on all datasets. Besides, both MAP of each attribute and the overall MAP are reported.

\subsubsection{Implementation Details}

For the backbones of both region-aware and patch-aware branches, we respectively choose a ResNet-50 network and ViT-B/16 network pre-trained on ImageNet.
For training strategy, we directly follow the two-stage training strategy used in \cite{dong2021fine}.
For hyper-parameters in \myeq{eq:total}, we empirically set $\beta=0.1$ and $\gamma=0.04$ to make all loss elements have a similar loss value at the beginning of the model training.
At the same time, the temperature factor $\tau$ and the larger penalty factor $\mu$ in \myeq{eq:bec} are empirically set to 0.07 and 12 respectively.
Note that before training, we construct 100k triplets on each dataset for model learning unless otherwise stated.
The model that performs the best on the validation set is used for evaluation on the test set.
Our source code is available at \url{https://github.com/HuiGuanLab/RPF}.

\subsection{Comparison to the State-of-the-art}\label{ssec:sota}
Table \ref{tab:sota-fashinai} summarizes the performance of different models on FashionAI. Besides the previous state-of-the-art methods, we also compare our model to a triplet network baseline. The baseline learns an attribute-agnostic embedding space, which directly employs mean pooling on the feature map generated by CNN without considering the attribute, and a standard triplet ranking loss is used to train the model.
Unsurprisingly, the baseline that learns an attribute-agnostic embedding space is much worse than other methods that learn attribute-aware embedding spaces. 
The results show the benefit of learning attribute-aware embedding spaces for ASFR.
Among the attribute-aware models, the majority of models are of a single branch architecture, \ie CSN \cite{veit2017csn}, ASEN \cite{ma2020fine}, HAEN \cite{yan2021learning}, AttnFashion\cite{wan2022learning}, ISLN\cite{yan2022attribute}.
As shown in this table, our proposed RPF of two branch architecture outperforms their performance by a clear margin. On this dataset, our RPF achieves 70.11 overall MAP while the best previous method ISLN~\cite{yan2022attribute} only hits 66.10. 
Especially on the attributes that are related to small regions, RPF outperforms the previous best performer, \ie ISLN~\cite{yan2022attribute}, by 14.3\% relatively on the \textit{neck design} attribute and 15.3\% relatively on the \textit{sleeve length} attribute.
It not only demonstrates the effectiveness of our two-branch solution, but also shows its superiority in the attributes that are related to small regions.
Additionally, ASEN$^{++}$ \cite{dong2021fine} is also a two-branch network, but their two branches are at the same granularity for modeling. By contrast, our proposed RPF is of a coarse-to-fine two-branch network and consistently gives better performance.
Table \ref{tab:sota-darn} and Table \ref{tab:sota-deepfashion} summarize the results on DARN and DeepFashion respectively.
Likewise, our proposed RPF model consistently outperforms all the other models on all attributes.
The results further verify the effectiveness of the proposed two-branch RPF model for ASFR.

\input{table-ablation-crossfashionai}
\input{table-ablation-structure}
\subsection{Generalization on Out-of-domain Data}
The above experiments are generally trained and evaluated on the same dataset where images are from the same domain.
However, it is important for models to generalize to out-of-domain data, especially in real application scenarios.
In this experiment, we explore the generalization of models by cross-dataset evaluation. Concretely, we first train a model on one dataset, and then evaluate its performance on the other dataset constructed in different ways.
Besides, we compare two models, \ie ASEN and $\text{ASEN}^{++}$, considering they are the only two works that have released the source code.

Table \ref{tab:crossdata} shows the cross-dataset evaluation results on FashionAI $\rightarrow$ DARN and DARN $\rightarrow$ FashionAI. 
Compared to ASEN of one branch, ASEN++ with two branches though achieve better performance for in-domain evaluation as shown in Table  \ref{tab:sota-fashinai} and Table \ref{tab:sota-darn}, it does not improve the generalization ability of the model.
By contrast, our proposed RPF not only achieves the best on the in-domain evaluation on three datasets, but also outperforms the counterparts with clear margins on cross-dataset evaluation.
We speculate it to our carefully designed coarse-to-fine two-branch architecture with foreground-background contrastive learning, which allows the model to learn fine-grained representation at various granularities thus improving the generalization on out-of-domain images.
\input{eval/tsne}
\input{table-ablation-loss2}
\input{table-ablation-module}

\subsection{Ablation Studies}\label{ssec:ablation}
To verify the effectiveness of each component in our proposed model, we conduct ablation studies on FashionAI.
For the experiments of ablation studies, we construct 10k triplets instead of 100k to reduce the training time of each variant.

\subsubsection{The Effectiveness of Region to Patch Framework} \label{section:rpf}
In order to verify the effectiveness of our proposed coarse-to-fine two-branch framework, \ie \textit{Region to Patch}, we compare the one-branch and two-branch solutions without the coarse-to-fine manner.
As the one-branch framework can not utilize inter-branch contrastive learning, all frameworks utilize intra-branch contrastive learning for a fair comparison.
The results are summarized in Table \ref{tab:ablation-structrue}.
It is worth noting that the \textit{Region} indicates inputting images in a whole region and utilizing CNN as the backbone, while the \textit{Patch} indicates inputting images in a form of patches and employing ViT as the backbone. 
The two one-branch frameworks, \ie \textit{Region Only} and \textit{Patch Only}, are worse than the other three frameworks of the two-branch. It demonstrates the advantage of employing two branches to extract fine-grained features.
Among the three two-branch ones, our proposed Region to Patch framework in a coarse-to-fine manner gives the best performance, which verifies the effectiveness of our framework for ASFR.

In addition, we also visualize the obtained attribute-specific embeddings of all models by t-SNE.
The results on all test images of FashionAI are illustrated in Figure \ref{fig:attr_tnse}, where the same color indicates the same attribute.
Dots of the same color obtained by our proposed framework are more clustered than that obtained by others.
This visualization demonstrates the better discriminatory ability of the learned attribute-specific embeddings by our proposed RPF.

\subsubsection{The Effectiveness of Intra-branch and Inter-branch Contrastive Learning}
In this section, we compare our full model to the degenerated counterparts with intra-branch or inter-branch contrastive learning only.
As shown in Table \ref{tab:ablation-IICL}, the one without the inter-branch contrastive learning is much worse than the other two with the inter-branch, which demonstrates the necessity of inter-branch contrastive learning for ASFR.
On the basis of inter-branch contrastive learning, including intra-branch contrastive learning achieves significant performance gain. 
It not only shows the complementarity of two contrastive learning ways, but also verifies the effectiveness of their joint use.

\input{table-ablation-AT.tex}
\input{eval/alpha}
\input{eval/example_multiattri}

\input{eval/example_attention}

\subsubsection{The Effectiveness of E-infoNCE loss}
In order to further explore the effect of positive and negative samples in E-infoNCE loss, we compare four kinds of combinations in Table \ref{tab:ablation-pn}.
Including extra positive samples of the foreground representation or the negative samples of the background representation in the mini-batch consistently achieve performance gain. Besides, including extra positive samples is more beneficial. Jointly using both extra positive samples and negative samples gives the best performance, which demonstrates the effectiveness of our proposed E-infoNCE.

\subsubsection{The Effectiveness of attribute-aware Transformer}
As shown in Table \ref{tab:ablation-module}, removing the attribute-aware Transformer module from the patch-aware branch  results in significant performance degradation.
It demonstrates the effectiveness of the attribute-aware Transformer module in the patch-aware branch for ASFR.

\subsubsection{The Influence of Hyper-parameters in RPF}
The influence of $\lambda$ in \myeq{lambda} is shown in Figure \ref{fig:lambda}.
We change $\lambda$ with its value ranging from 0 to 1 with an interval of 0.1, and the performance of RPF reaches its peak with an $\lambda$ of 0.3.
The result indicates that our proposed RPF more rely on the patch-aware branch, which is due to much better performance of the patch-aware branch than the region-aware one.
Thus, a small $\lambda$ is suggested for better performance.
Additionally, the influence of $\mu$ in \myeq{eq:bec} is shown in Figure \ref{fig:alpha}.
The performance of our RPF reaches its peak with an $\mu$ of 12, and it can be observed that our RPF is not very sensitive to hyper-parameter $\mu$.

\subsection{Visualization Analysis}
\subsubsection{Retrieval Examples}
Figure \ref{fig:multi} illustrates several fine-grained attribute-specific fashion retrieved results obtained by our proposed RPF model.
It is obvious that most of the retrieved images share the same specified attributes with the query image.
Taking the attribute \textit{lapel design} as an example, the retrieved images are of \textit{notch lapel} with the same lapel design as the query image.
These results demonstrate that RPF is expert in capturing attribute-related representations in fashion items.

Moreover, our RPF also allows for multi-attribute composition retrieval. 
Specifically, the similarities of given multiple attributes are summed up as the final similarity score, which is used to rank candidate images.
As illustrated in the second row of Figure \ref{fig:multi}, given two attributes, \textit{lapel design} and \textit{collar design}, the retrieved results are obviously the same in terms of the composite attributes. 
\subsubsection{Attention Visualization}
To further explore the capability of region-aware and patch-aware branches for locating attribute-related regions, we respectively visualize the attention weights from both branches.
As shown in Figure \ref{fig:attention}, the region-aware branch usually can locate the regions roughly corresponding to the given attribute but the regions are not precise enough.
By contrast, as the patch-aware branch is at a more fine-grained level and the attribute-aware Transformer can further filter out unrelated patches, the localization of this branch has higher accuracy.
By combining region-aware and patch-aware branches in a coarse-to-fine manner, our model can finally achieve superior retrieval results.

%% file: table-sota-deepfashion.tex
\begin{table}[tb!]
\renewcommand{\arraystretch}{1.}
\caption{Performance comparison on DeepFashion. 
}
\vspace{-3mm}
\label{tab:sota-deepfashion}
\centering 
\scalebox{0.85}{
\begin{tabular}{l*{6}{c}}
\toprule
\multirow{2}{*}{\textbf{Method}} & \multicolumn{5}{c}{\textbf{MAP for each attribute}} & \multirow{2}{*}{\textbf{overall MAP}} \\
\cmidrule(l){2-6}
 &texture&fabric&shape&part&style \\
\cmidrule(l){1-7}
Triplet baseline & 13.26 & 6.28 & 9.49 & 4.43 & 3.33 & 7.36 \\
CSN \cite{veit2017csn}                & 14.09 & 6.39 & 11.07 & 5.13 & 3.49 & 8.01 \\
AttnFashion \cite{wan2022learning}        & 12.90 & 6.34 & 11.38 & 5.24 & 4.20 & 8.01 \\
$\text{ASEN}$ \cite{ma2020fine}    & 15.01 & 7.32 & 13.32 & 6.27 & 3.85 & 9.14 \\
$\text{ASEN}^{++}$ \cite{dong2021fine}    & 15.60 & 7.67 & 14.31 & 6.60 & 4.07 & 9.64 \\
RPF (ours)        & \textbf{15.62} & \textbf{8.30} & \textbf{15.02} & \textbf{7.38} & \textbf{4.77} & \textbf{10.22}  \\
\bottomrule
\end{tabular}
}
\vspace{-4mm}
\end{table}

%% file: table-ablation-crossfashionai.tex

\begin{table} [tb!]
\renewcommand{\arraystretch}{1.}
\caption{Cross-dataset evaluation on FashionAI $\rightarrow$ DARN and DARN $\rightarrow$ FashionAI. $A\rightarrow B$ denotes the setting of training on $A$ dataset and evaluation on $B$ dataset. Our proposed RPF shows better generalization on out-of-domain data.}
\vspace{-3mm}
\label{tab:crossdata}
\centering 
\scalebox{1.0}{
\begin{tabular}{l*{4}{c}}
\toprule
\multirow{2}{*}{\textbf{Method}} &\multicolumn{3}{c}{$\textbf{FashionAI} \rightarrow \textbf{DARN}$} \\
\cmidrule(l){2-4}
&sleeve length &clothes length &collar shape  \\
\cmidrule(l){1-4}
$\text{ASEN}$~\cite{ma2020fine}     & 70.63 & 44.10& 24.36   \\
$\text{ASEN}^{++}$~\cite{dong2021fine} & 70.68 & 44.55 & 24.39  \\
RPF (ours)       &\textbf{71.04} &\textbf{45.53} &\textbf{24.83} \\
\cmidrule(l){1-4}
\multirow{2}{*}{\textbf{Method}} &\multicolumn{3}{c}{$\textbf{DARN} \rightarrow \textbf{FashionAI}$} \\
\cmidrule(l){2-4}
 &sleeve length &coat length &neckline design \\
\cmidrule(l){1-4}
$\text{ASEN}$~\cite{ma2020fine}      &29.64 & 25.37 & 17.15 \\
$\text{ASEN}^{++}$~\cite{dong2021fine}  &31.05 & 26.57 & 17.61\\
RPF (ours)    &\textbf{35.03} &\textbf{28.94} &\textbf{21.16}\\
\bottomrule
\end{tabular}
}
\vspace{-3mm}
\end{table}

%% file: table-ablation-structure.tex
\begin{table*} [ht]
\renewcommand{\arraystretch}{1.}
\caption{Performance comparison of different frameworks. Our proposed \textit{Region to Patch} framework of a coarse-to-fine manner significantly outperforms the other counterparts.}
\label{tab:ablation-structrue}
\vspace{-3mm}
\centering 
\scalebox{0.87}{
\begin{tabular}{l*{9}{c}}
\toprule
\multirow{2}{*}{\textbf{Architecture}} & \multicolumn{8}{c}{\textbf{MAP for each attribute}} & \multirow{2}{*}{\textbf{overall MAP}} \\
\cmidrule(l){2-9}
& skirt length & sleeve length & coat length & pant length & collar design & lapel design & neckline design & neck design \\
\cmidrule(l){1-10}
Region Only                 & 64.44 & 54.63 & 51.27 & 63.53 & 70.79 & 65.36 & 59.50 & 58.67 & 61.02\\
Patch Only    & 65.23 & 57.07 & 58.11 & 70.81 & 62.27 & 58.15 & 56.57 & 53.30 & 60.00 \\
Region to Region      & 65.58 & 56.22 & 53.19 & 68.16 & 75.38 & 69.92 & 65.58 & 68.61 & 63.92 \\
Patch to Patch   & 64.96 & 60.03 & 57.72 & 70.35 & 68.14 & 65.21 & 62.85 & 57.98  & 63.06 \\
Region to Patch & \textbf{68.15} & \textbf{64.68} & \textbf{58.67} & \textbf{73.59} & \textbf{78.03} & \textbf{71.05} & \textbf{74.09} & \textbf{75.30} & \textbf{69.63}\\
\bottomrule
\end{tabular}
}
\vspace{-2mm}
\end{table*}

%% file: eval/tsne.tex
\begin{figure*}
	\centering
    \subcaptionbox{Region Only}
	{\includegraphics[width=0.15\textwidth]{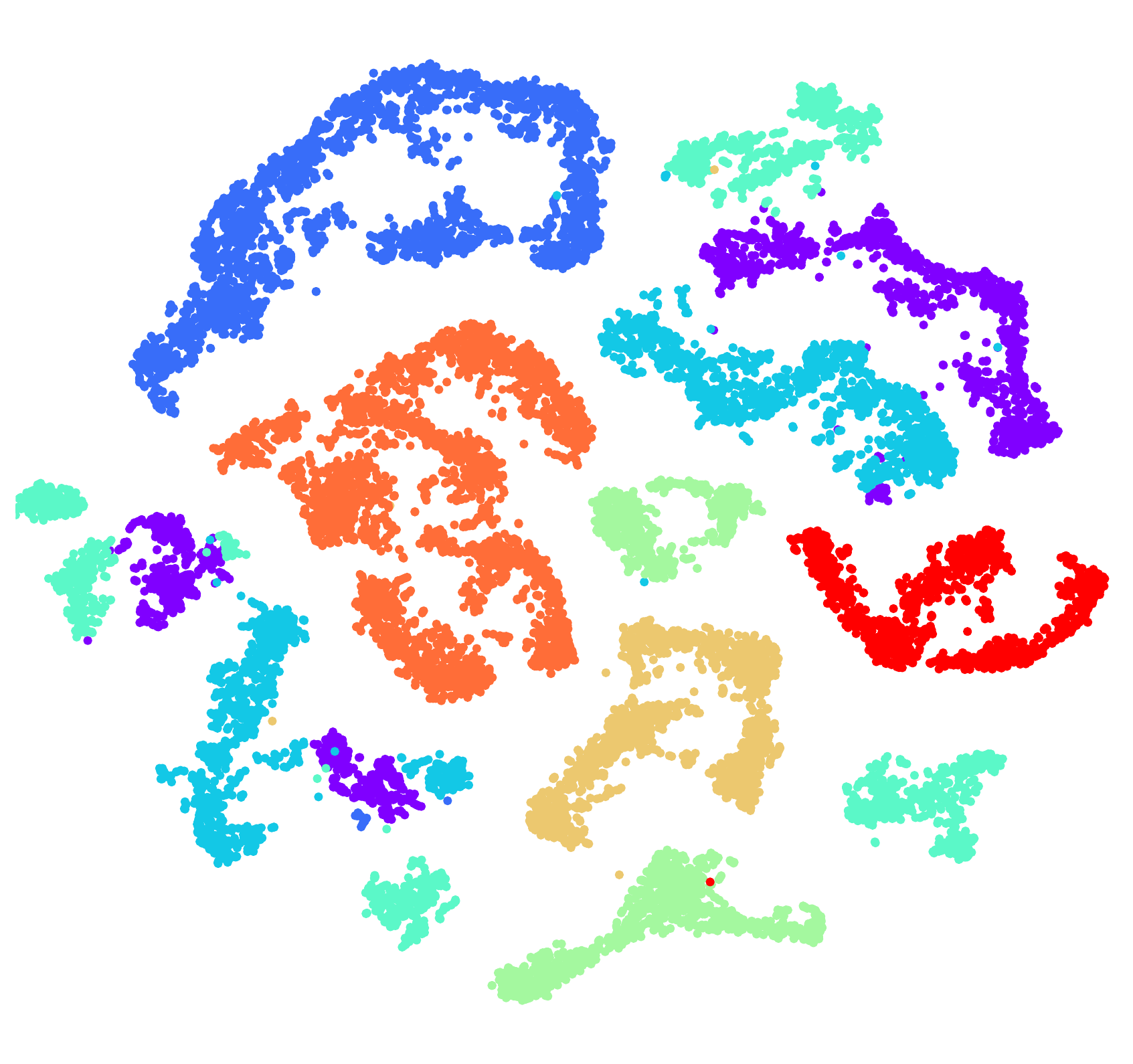}}
    \hfill
    \subcaptionbox{Patch Only}
	{\includegraphics[width=0.15\textwidth]{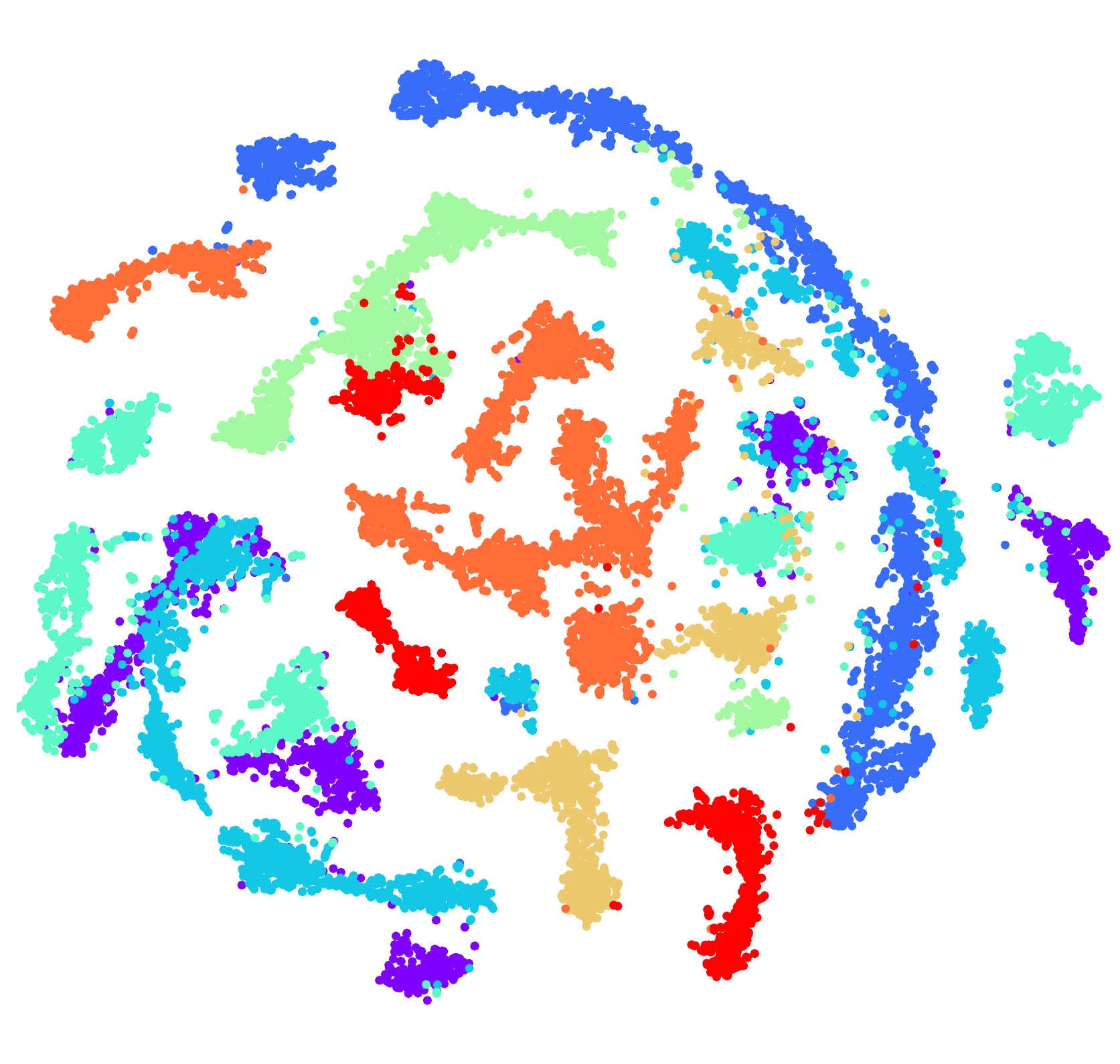}}
    \hfill
	\subcaptionbox{Region to Region}
	{\includegraphics[width=0.15\textwidth]{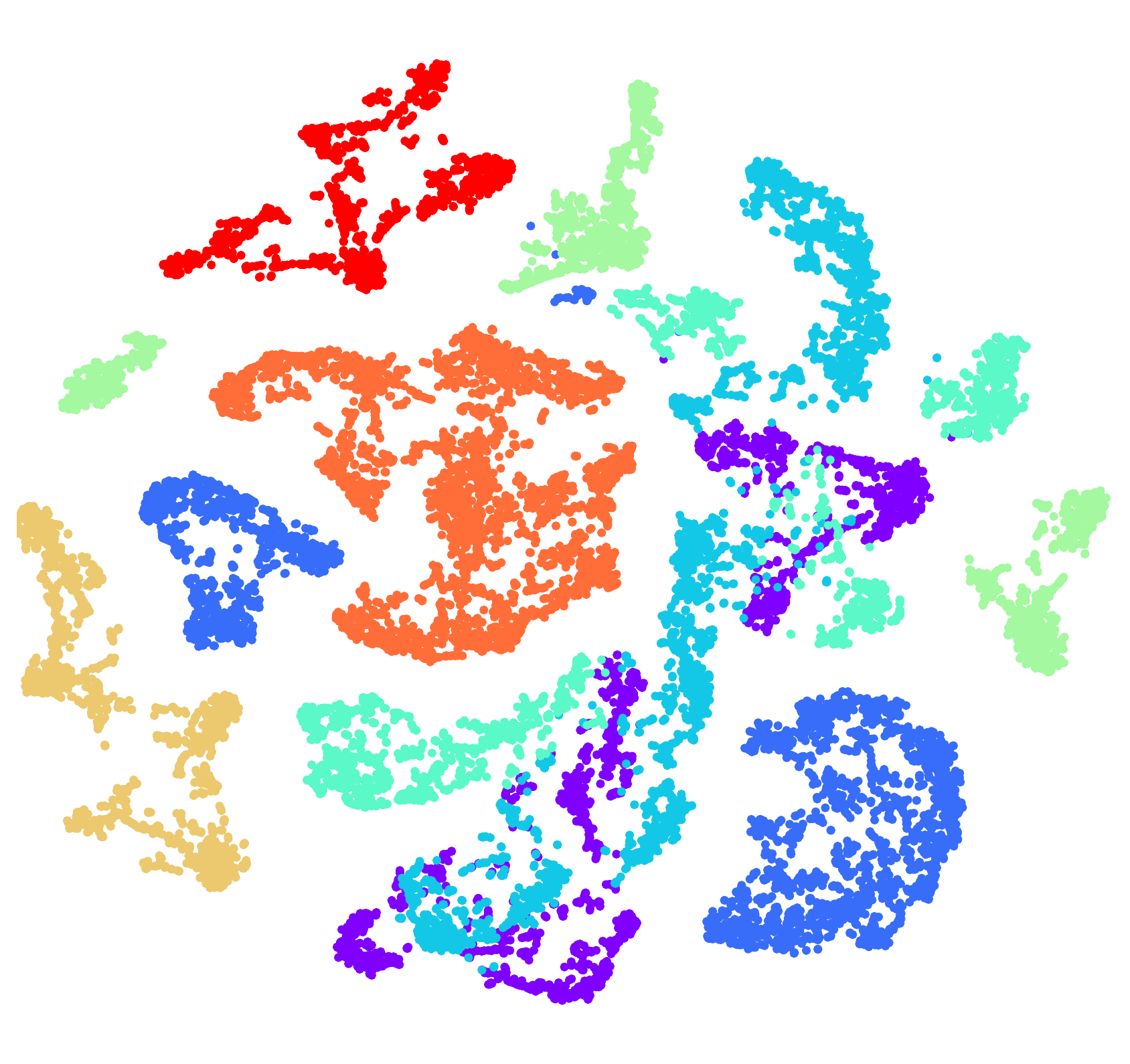}}
    \hfill
    \subcaptionbox{Patch to Patch}
	{\includegraphics[width=0.15\textwidth]{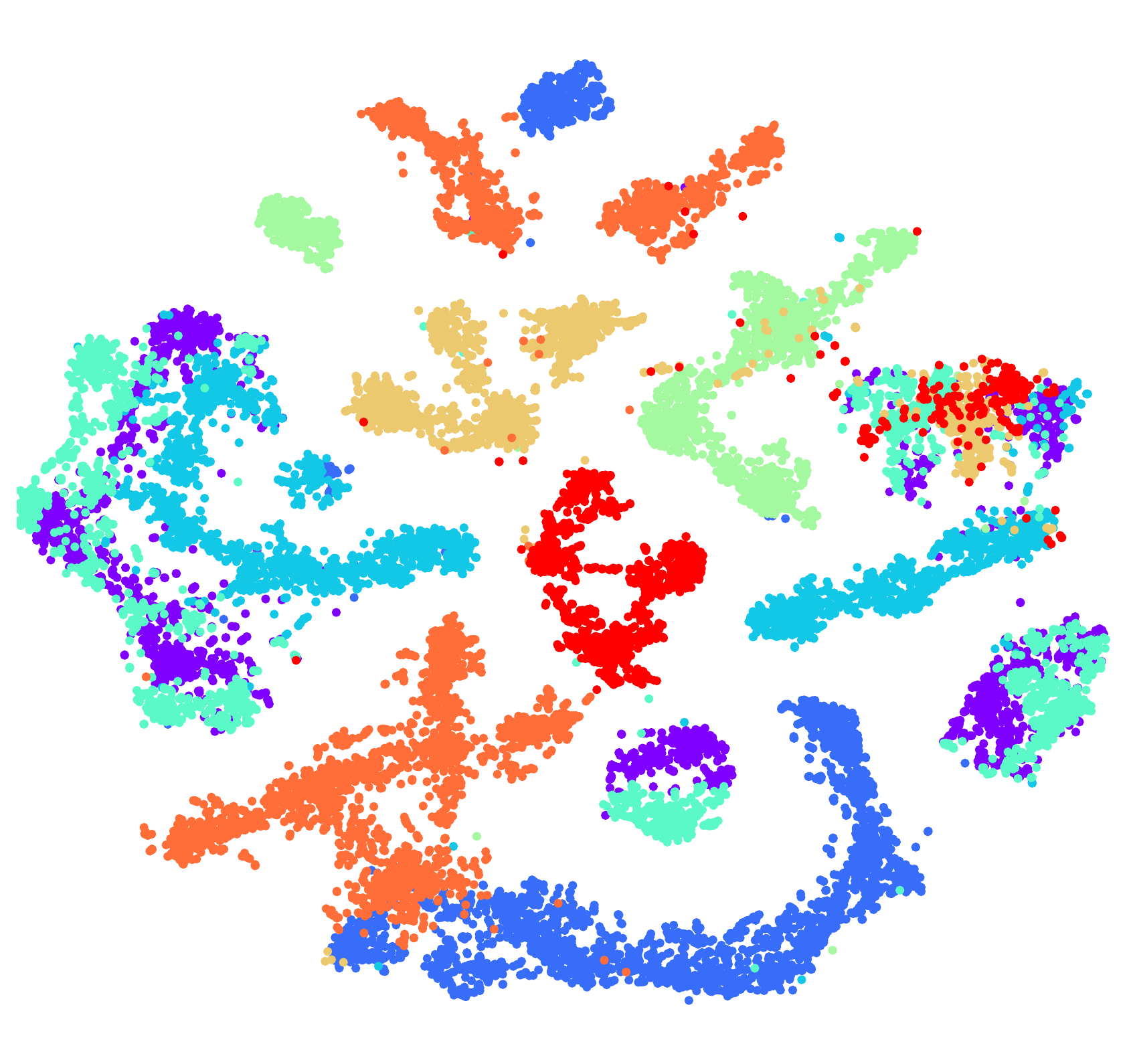}}
	\hfill
    \subcaptionbox{Region to Patch}
	{\includegraphics[width=0.22\textwidth]{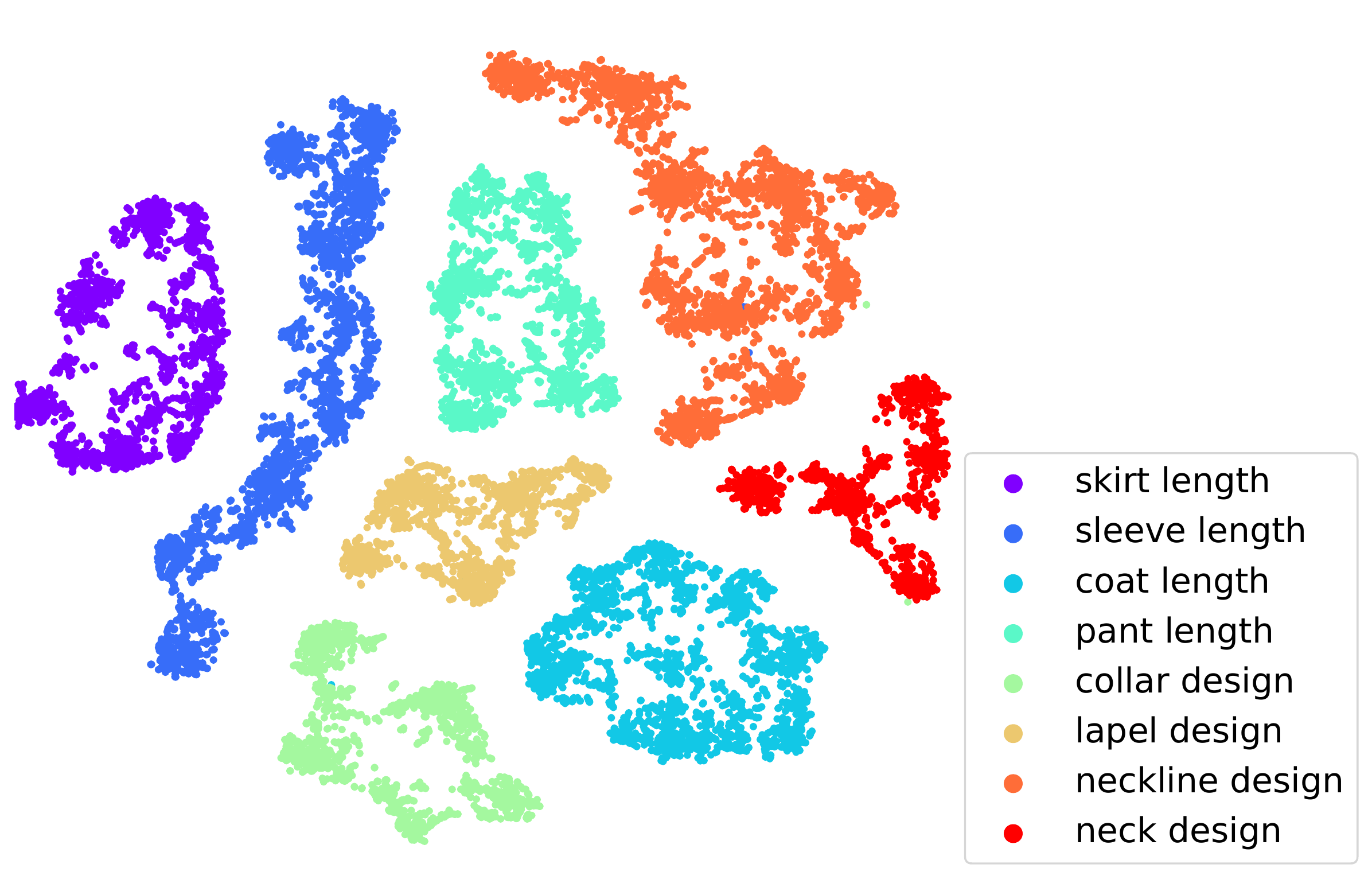}}
 \vspace{-4mm}
	\caption{T-SNE visualization of embedding spaces learned by different frameworks.
Dots with the same color indicate images having the same attribute.}\label{fig:attr_tnse}
 \vspace{-3mm}
\end{figure*}


%% file: table-ablation-loss2.tex
\begin{table}[ht]
\renewcommand{\arraystretch}{1.}
\caption{Ablation study on foreground-background contrastive learning.
\label{tab:ablation-IICL}}
\vspace{-3mm}
\centering 
\scalebox{1}{
\begin{tabular}{*{1}l*{3}c}
\toprule
&\textbf{Intra-branch} & \textbf{Inter-branch} & \textbf{overall MAP}\\
\cmidrule{0-3}
&\ding{55} &\checkmark &55.39  \\ 
&\checkmark &\ding{55} & 66.62 \\
&\checkmark &\checkmark & \textbf{69.42} \\ 
\bottomrule
\end{tabular}}
\vspace{-1.5mm}
\end{table}

%% file: table-ablation-module.tex
\begin{table}[ht]
\renewcommand{\arraystretch}{1.}
\caption{The effect of positive and negative samples in E-infoNCE loss. Note that E-infoNCE without $\exp{( {r}_i\cdot{ {p}_j^+}/\tau)}$ and $\exp{( {r}_i\cdot{ \widetilde{{r}_j}}/\tau)}$ degenerates into the normal infoNCE. 
}
\vspace{-3mm}
\label{tab:ablation-pn}
\centering 
\scalebox{0.9}{
\begin{tabular}{*{1}l*{3}c}
\toprule
& \textbf{positive:} $\exp{( {r}_i\cdot{ {p}_j^+}/\tau)}$ &  \textbf{negative:} $\exp{( {r}_i\cdot{ \widetilde{{r}_j}}/\tau)}$ & \textbf{overall MAP}\\
\cmidrule{1-4}
 &\ding{55}  &\ding{55} &65.42 \\
 &\ding{55}  &\checkmark &66.78 \\
 &\checkmark &\ding{55} &68.99 \\
 &\checkmark &\checkmark &69.42 \\
\bottomrule
\end{tabular}
}
\vspace{-3mm}
\end{table}

%% file: table-ablation-AT.tex
\begin{table}[tb]
\renewcommand{\arraystretch}{1.}
\caption{The Effectiveness of attribute-aware Transformer. 
}
\vspace{-3mm}
\label{tab:ablation-module}
\centering 
\scalebox{1}{
\begin{tabular}{*{1}l*{2}c}
\toprule
&\textbf{attribute-aware Transformer}  & \textbf{overall MAP}\\
\cmidrule{0-2}
&\ding{55}  & 66.61 \\ 
&\checkmark  & \textbf{69.42} \\ 
\bottomrule
\end{tabular}}
\vspace{-1mm}
\end{table}

%% file: eval/alpha.tex
\begin{figure}[tb!]
	\centering
	\subcaptionbox{\label{fig:lambda}}
	{\includegraphics[width=0.49\linewidth]{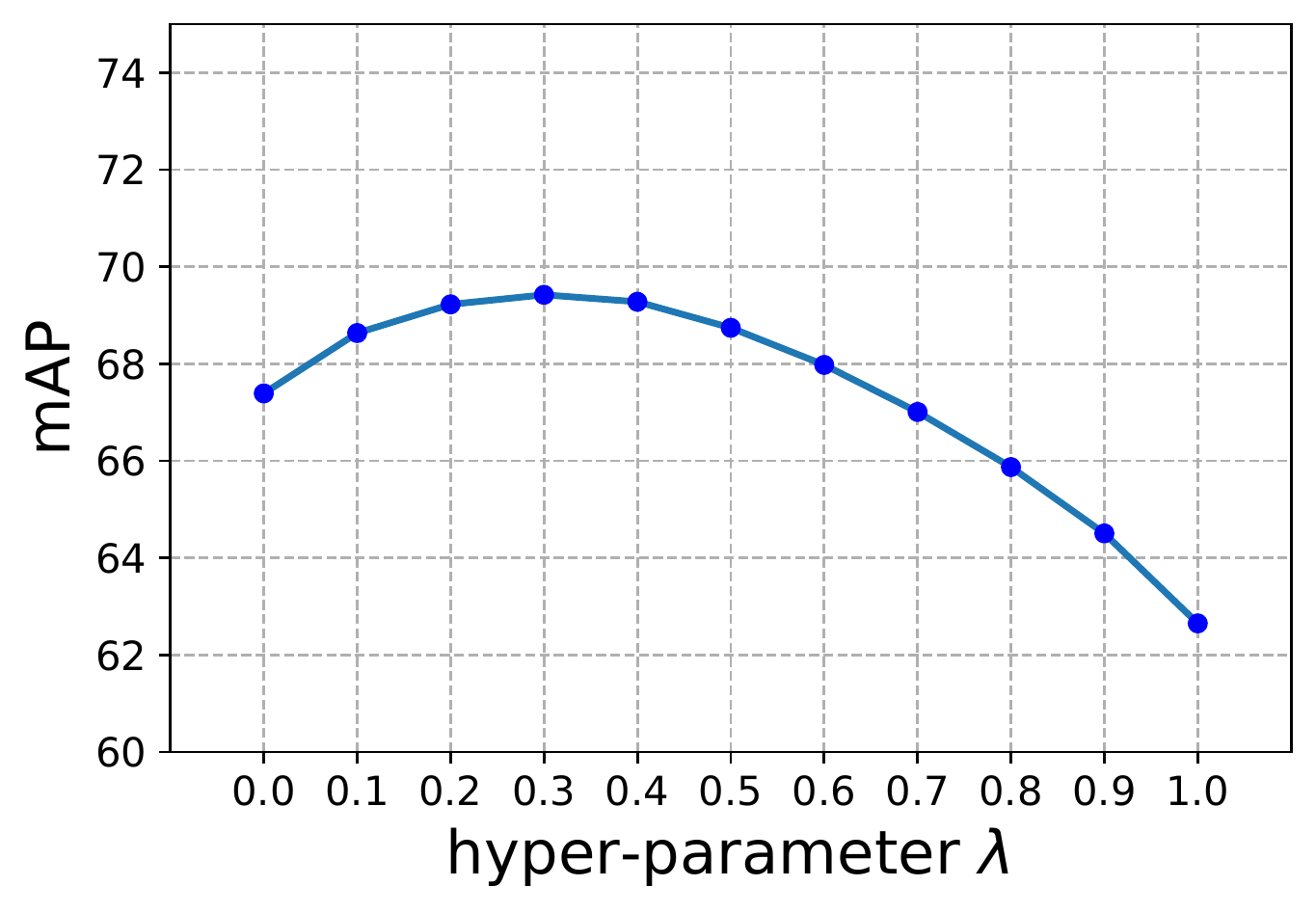}}
	\subcaptionbox{\label{fig:alpha}}
	{\includegraphics[width=0.49\linewidth]{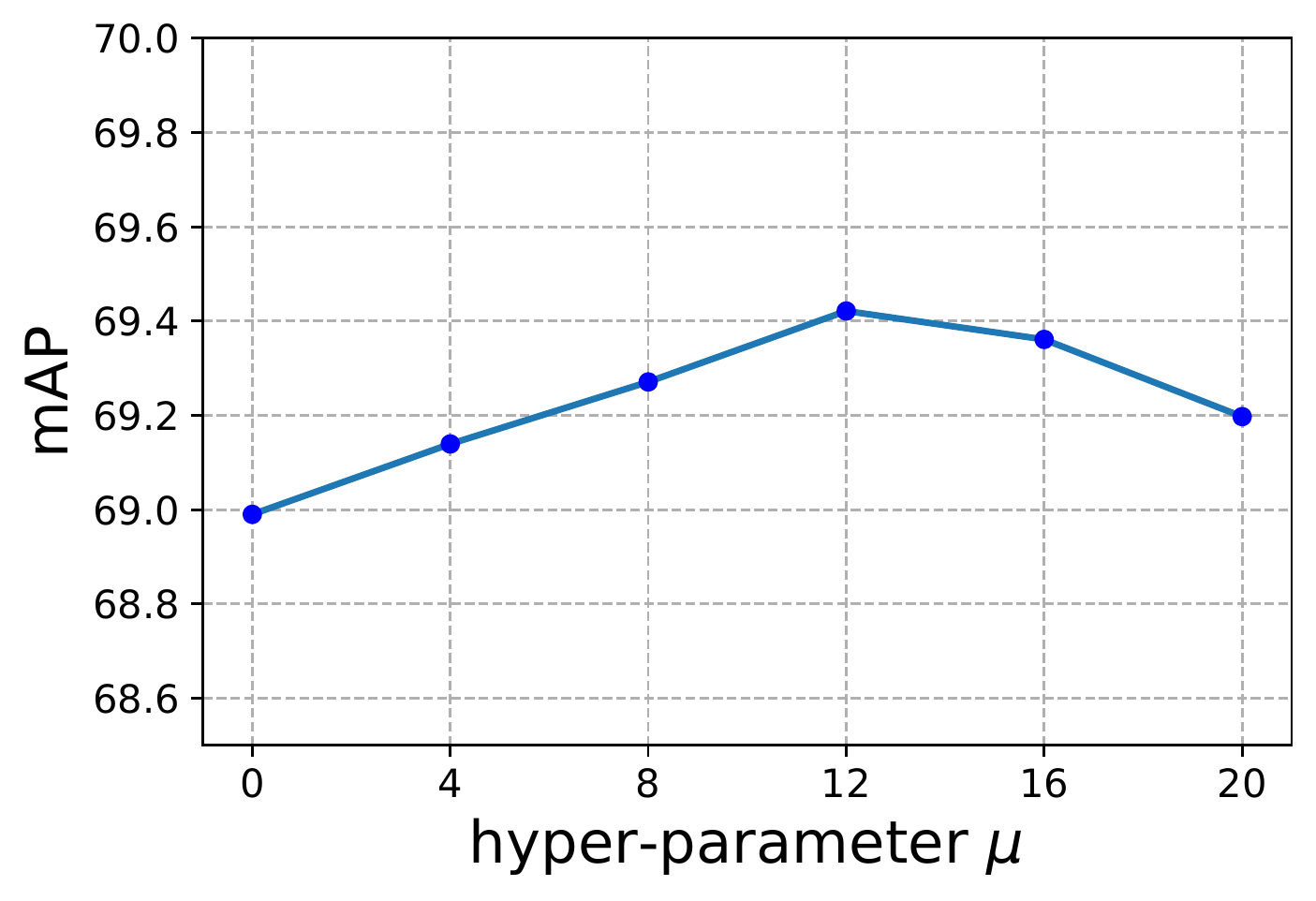}}
\vspace{-3mm}	
 \caption{The influence of (a) hyper-parameter $\lambda$ of \myeq{lambda} and (b) hyper-parameter $\mu$ of \myeq{eq:bec} in RPF on FashionAI.}
\vspace{-3mm}
\end{figure}

%% file: eval/example_multiattri.tex
\begin{figure*}[ht]
	\centering
	\label{fig:multiv2}
	\includegraphics[width=.97\linewidth]{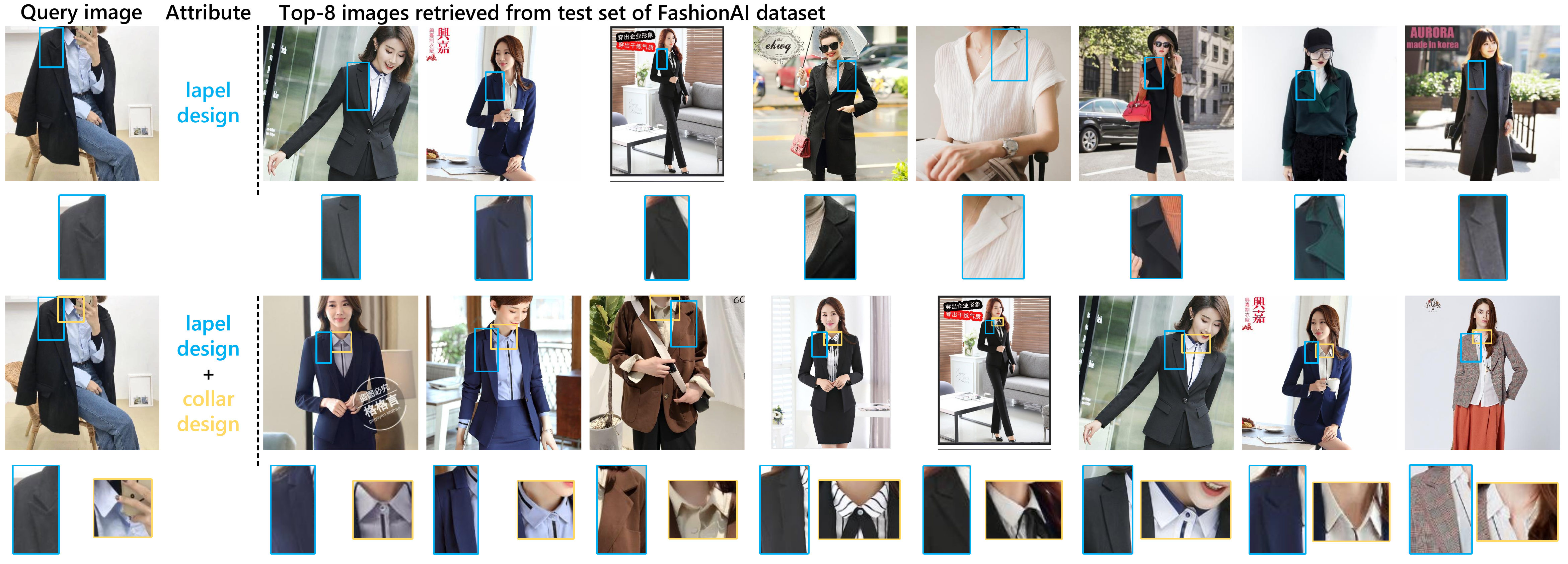}
\vspace{-5mm}	
 \caption{Attribute-specific fashion retrieval examples on FashionAI. Given a query image and a specified attribute, our model is able to retrieve images having the corresponding same attribute value as the query image.  
 Besides, as exemplified in the second row, our proposed RPF allows for multi-attribute composition retrieval, which supports for \textit{multiple} attributes as the input to search images of the same values for the multiple input attributes.
 \label{fig:multi}}
 \vspace{-3mm}	
\end{figure*}

%% file: eval/example_attention.tex
\begin{figure}[ht]
\centering
\includegraphics[width=.96\columnwidth]{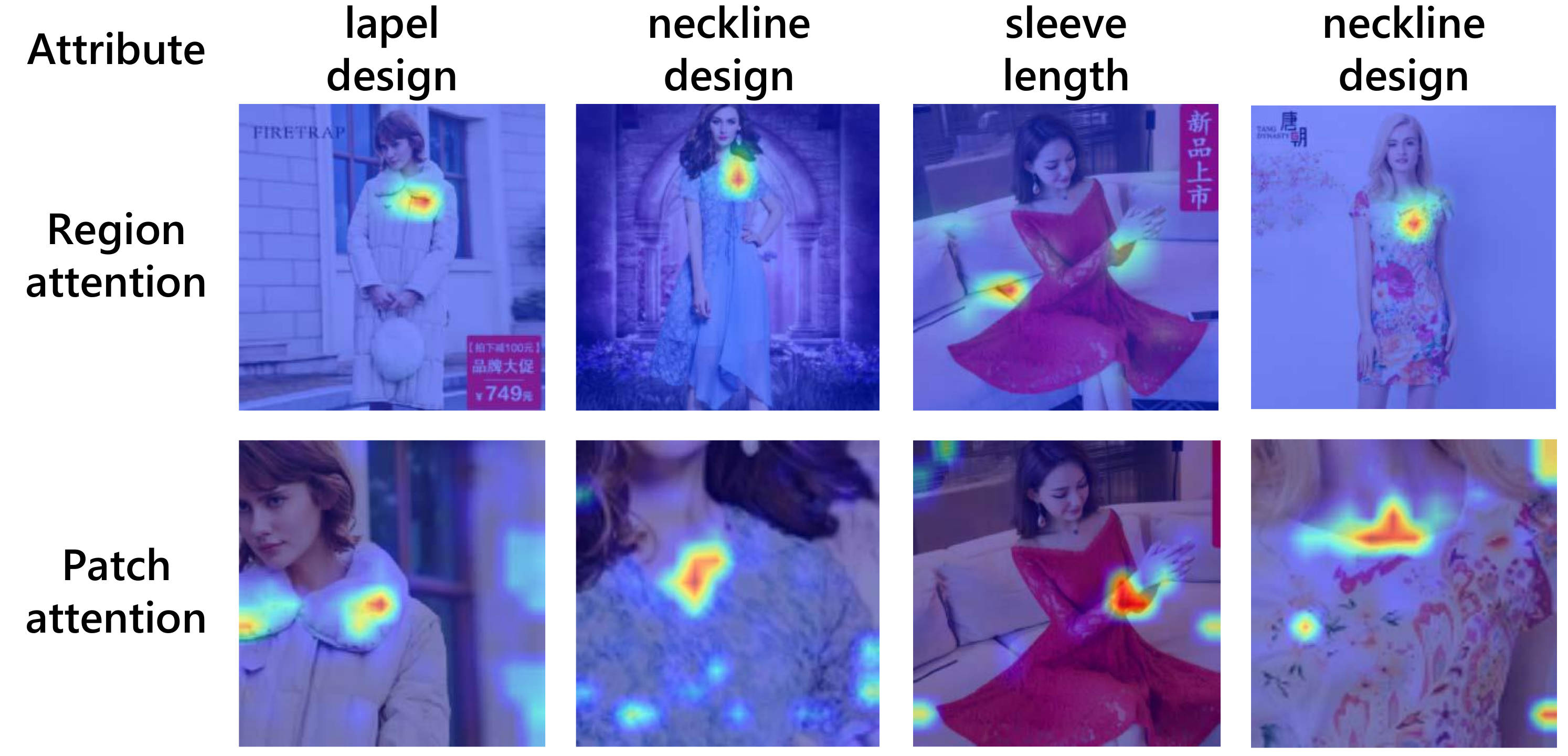}
\vspace{-4mm}
\caption{Visualization of the attribute-aware attention weights obtained by the region-aware branch (first row) and the patch-aware branch (second row).
}
\vspace{-3mm}
\label{fig:attention}
\end{figure}